\documentclass[11pt,a4paper]{article}
\usepackage{url}
\usepackage{placeins}
\usepackage[printwatermark]{xwatermark}
\usepackage{xcolor}
\usepackage{graphicx}

\usepackage[hyperref]{naaclhlt2019}
\usepackage{times}
\usepackage{latexsym}
\usepackage{latexsym}
\usepackage{graphicx}
\usepackage{todonotes}
\usepackage{microtype}
\usepackage{soul}
 \usepackage{array}
\usepackage{hyperref}
 \usepackage{colortbl, booktabs}
\usepackage{todonotes}
\graphicspath{{graphics/}}

\usepackage{multirow}

\aclfinalcopy 
\def\aclpaperid{1307} 

\def\rot{\rotatebox}
\title{
Training Data Augmentation for Context-Sensitive Neural Lemmatization\\ Using Inflection Tables and Raw Text
}

\author{Toms Bergmanis \\
  School of Informatics \\
  University of Edinburgh \\
  {\tt T.Bergmanis@sms.ed.ac.uk} \\\And
  Sharon Goldwater \\
  School of Informatics \\
  University of Edinburgh \\
  {\tt sgwater@inf.ed.ac.uk} \\}

\date{}

\begin{document}
\maketitle
\begin{abstract}
Lemmatization aims to reduce the sparse data problem by relating the inflected forms of a word to its dictionary form. Using context can help, both for unseen and ambiguous words. Yet most context-sensitive approaches require full lemma-annotated sentences for training, which may be scarce or unavailable in low-resource languages. In addition (as shown here), in a low-resource setting, a lemmatizer can learn more from $n$ labeled examples of distinct words (types) than from $n$ (contiguous) labeled tokens, since the latter contain far fewer distinct types. To combine the efficiency of type-based learning with the benefits of context, we propose a way to train a context-sensitive lemmatizer with little or no labeled corpus data, using inflection tables from the UniMorph project and raw text examples from Wikipedia that provide sentence contexts for the unambiguous UniMorph examples. Despite these being unambiguous examples, the model successfully generalizes from them, leading to improved results (both overall, and especially on unseen words) in comparison to a baseline that does not use context.

\end{abstract}

\section{Introduction}

Many lemmatizers work on isolated wordforms 
\citep{wicentowski2002modeling,dreyer2008latent,rastogi2016weighting,makarov2018neural,makarov2018imitation}. Lemmatizing in context can improve accuracy on ambiguous and unseen words \cite{bergmanis.goldwater:context}, but most systems for context-sensitive lemmatization must train on complete sentences labeled with POS and/or morphological tags as well as lemmas, and have only been tested with 20k-300k training tokens \citep{chrupala2008learning,muller2015joint,chakrabarty2017context}.\footnote{The smallest of these corpora contains 20k tokens of Bengali annotated only with lemmas, which \citet{chakrabarty2017context} reported took around two person months to create.}

Intuitively, though, sentence-annotated data is inefficient for training a lemmatizer, especially in low-resource settings. Training on (say) 1000 word types will provide far more information about a language's morphology than training on 1000 contiguous tokens, where fewer types are represented. As noted above, sentence data can help with ambiguous and unseen words, but we show here that when data is scarce, this effect is small relative to the benefit of seeing more word types.\footnote{\citet{garrette2013real} found the same for POS tagging.}
Motivated by this result, we propose a training data augmentation method that combines the efficiency of type-based learning and the expressive power of 
a context-sensitive model.\footnote{Code and data: \url{https://bitbucket.org/tomsbergmanis/data_augumentation_um_wiki}}
We use  \textit{Lematus} \cite{bergmanis.goldwater:context}, a state-of-the-art lemmatizer that learns from lemma-annotated words in their $N$-character contexts. No predictions about surrounding words are used, so fully annotated training sentences are not needed.
We exploit this fact by combining two sources of training data: 
1k lemma-annotated types (with contexts) from the Universal Dependency Treebank (UDT) v2.2\footnote{\url{http://hdl.handle.net/11234/1-2837}} \cite{UDT2}, plus examples obtained by finding \emph{unambiguous} word-lemma pairs in inflection tables from the Universal Morphology (UM) project\footnote{\url{http://unimorph.org}} and collecting sentence contexts for them from Wikipedia. Although these examples are noisy and biased, we show that they improve lemmatization accuracy in experiments on 10 languages, and that the use of context helps, both overall and especially on unseen words.

\section{Method}
\textbf{Lematus} \cite{bergmanis.goldwater:context} is a neural sequence-to-sequence model with attention inspired by the re-inflection model of \citet{kann2016med}, which won the 2016 SIGMORPHON shared task \cite{cotterell2016sigmorphon}. It is built using the Nematus machine translation toolkit,\footnote{Code for Nematus: \url{https://github.com/EdinburghNLP/
nematus}, Code for Lematus: \url{https://bitbucket.org/tomsbergmanis/lematus.git}} which uses the architecture of \newcite{DBLP:journals/corr/SennrichFCBHHJL17}: a 2-layer bidirectional GRU encoder and a 2-layer decoder with a conditional GRU \cite{DBLP:journals/corr/SennrichFCBHHJL17} in the first layer and a GRU in the second layer.

Lematus takes as input a character sequence representing the wordform in its $N$-character context, and outputs the characters of the lemma. Special input symbols are used to represent the left and right boundary of the target wordform (\texttt{<lc>}, \texttt{<rc>}) and other word boundaries (\texttt{<s>}). For example, if $N=15$, the system trained on Latvian would be expected to produce the characters of the lemma \textit{ce\c{l}\v{s}} (meaning \textit{road}) given input such as:
\begin{center}
\texttt{ s a k a <s> p a \v{s} v a l d \={\i} b u } \\
\texttt{<lc> c e \c{l} u <rc> } \\ 
\texttt{u n <s> i e l u <s> r e \'{g} i s t r}
\end{center}

When $N=0$ (\textbf{Lematus 0-ch}), no context is used, making Lematus 0-ch comparable to other systems that do not model context \cite{dreyer2008latent,rastogi2016weighting,makarov2018neural,makarov2018imitation}. In our experiments we use both Lematus 0-ch and \textbf{Lematus 20-ch} (20 characters of context), which was the best-performing system reported by \citet{bergmanis.goldwater:context}.

\subsection{Data Augmentation}
Our data augmentation method uses UM inflection tables and creates additional training examples by finding Wikipedia sentences that use the inflected wordforms in context, pairing them with their lemma as shown in the inflection table. However, we cannot use all the words in the tables because some of them are ambiguous: for example, Figure~\ref{table:inflections} shows that the form \texttt{ce\c{l}i} could be lemmatized either as \texttt{ce\c{l}\v{s}} or \texttt{celis}.  Since we don't know which would be correct for any particular Wikipedia example, we only collect examples for forms which are unambiguous according to the UM tables.
However, this method is only as good as the coverage of the UM tables. For example, if UM doesn't include a table for the Latvian verb \emph{celt}, then the underlined forms in Table~\ref{table:inflections}  would be incorrectly labeled as unambiguous.

\begin{table}
\begin{center}

\begin{tabular}[t]{r|cc|cc}
& \multicolumn{2}{|c|}{\textbf{noun: \texttt{ce\c{l}\v{s}}}} &  \multicolumn{2}{|c}{\bfseries noun: \texttt{celis} }\\
& \textbf{\texttt{SG}}  & \textbf{\texttt{PL}} & \textbf{\texttt{SG}}  & \textbf{\texttt{PL}} \\ \hline
\texttt{NOM} & \texttt{ce\c{l}\v{s}}    & \st{\texttt{ce\c{l}i}} & \texttt{celis} & \st{\texttt{ce\c{l}i}}\\
\texttt{GEN} & \st{\texttt{ce\c{l}a}}    & \st{\texttt{ce\c{l}u}}  &  \st{\texttt{ce\c{l}a}}    & \st{\texttt{ce\c{l}u}} \\
\texttt{DAT}  & \underline{\texttt{ce\c{l}am}}& \st{\texttt{ce\c{l}iem}} & \texttt{celim}   & \st{\texttt{ce\c{l}iem}} \\
\texttt{ACC}  & \underline{\texttt{ce\c{l}u}}    & \st{\texttt{ce\c{l}us}}  & \texttt{celi} & \st{\texttt{ce\c{l}us}}  \\
\texttt{INS} & \underline{\texttt{ce\c{l}u}}   & \st{\texttt{ce\c{l}iem}} & \texttt{celi} & \st{\texttt{ce\c{l}iem}}\\
\texttt{LOC}  & \texttt{ce\c{l}\={a}}   & \st{\texttt{ce\c{l}os}} & \texttt{cel\={\i}} & \st{\texttt{ce\c{l}os}} \\
\texttt{VOC} & \underline{\texttt{ce\c{l}}}     & \st{\texttt{ce\c{l}i}}  & \texttt{celi} & \st{\texttt{ce\c{l}i}}
\end{tabular}

\caption{Example UM inflection tables for Latvian nouns \textit{ce\c{l}\v{s}} (\textit{road}) and \textit{celis} (\textit{knee}). The \st{crossed out forms} are examples of evidently ambiguous forms that are not used for data augmentation because of being shared by the two lemmas. The \underline{underlined forms} appear unambiguous in this toy example but actually conflict with inflections of the verb \textit{celt} (\textit{to lift}). \vspace{-12pt} }
\label{table:inflections}
\end{center}
\end{table}

There are several other issues with this method that could potentially limit its usefulness. First, the UM tables only include verbs, nouns and adjectives, whereas we test the system on UDT data, which includes all parts of speech. Second, by excluding ambiguous forms, we may be restricting the added examples to a non-representative subset of the potential inflections, or the system may simply ignore the context because it isn't needed for these examples. Finally, there are some annotation differences between UM and UDT.\footnote{Recent efforts to unify the two resources have mostly focused on validating dataset schema \citep{mccarthy2018marrying}, leaving conflicts in word lemmas unresolved. We estimated (by counting types that are unambiguous in each dataset but have different lemmas across them) that annotation inconsistencies affect up to 1\% of types in the languages we used.} Despite all of these issues, however, we show below that the added examples and their contexts do actually help.

\section{Experimental Setup}
\paragraph{Baselines and Training Parameters}
We use four baselines:
(1) \textbf{Lemming}\footnote{\url{http://cistern.cis.lmu.de/lemming}} \citep{muller2015joint} is a context-sensitive system that
uses log-linear models to jointly tag and lemmatize the data, and 
is trained on sentences annotated with both lemmas and POS tags. 
(2) The hard monotonic attention model (\textbf{HMAM})\footnote{\url{https://github.com/ZurichNLP/coling2018-  neural-transition-based-morphology}} \cite{makarov2018neural}
is a neural sequence-to-sequence model with a hard attention mechanism that advances through the sequence monotonically. It is trained on word-lemma pairs (without context) with character-level alignments learned in a preprocessing step using an alignment model, and it 
has proved to be competitive in low resource scenarios. 
(3) Our naive \textbf{Baseline} 
outputs the most frequent lemma (or one lemma at random from the options that are equally frequent) for words observed in training. For unseen words it outputs the wordform itself.
(4) We also try a baseline data augmentation approach (\textbf{AE Aug Baseline}) inspired by \citet{bergmanis-2017} and \citet{kann2017unlabeled}, who showed that adding training examples where the network simply learns to auto-encode corpus words can improve morphological inflection results in low-resource settings. The AE Aug Baseline is a variant of Lematus 0-ch which augments the UDT lemmatization examples by auto-encoding the inflected forms of the UM examples (i.e., it just treats them as corpus words). Comparing AE Aug Baseline to Lematus 0-ch augmented with UM lemma-inflection examples tells us whether using the UM lemma information helps more than simply auto-encoding more inflected examples.

To train the models we use the default settings for Lemming and the suggested lemmatization parameters for HMAM. We mainly follow the hyperparameters used by \citet{bergmanis.goldwater:context} for Lematus; details are in Appendix \ref{appendix:lemmatus_training}.

\paragraph{Languages and Training Data}
We conduct preliminary experiments on five development languages: Estonian,	Finnish, Latvian, Polish, and	Russian. In our final experiments we also add Bulgarian, Czech, Romanian, Swedish and Turkish.
We vary the amount and type of training data (types vs. tokens, UDT only, UM only, or UDT plus up to 10k UM examples), as described in Section~\ref{sec:results}.

To obtain $N$ UM-based training examples, we select the first $N$ unambiguous UM types (with their sentence contexts) from shuffled Wikipedia sentences.
For experiments with $j>1$ examples per type, we first find all UM types with at least $j$ sentence contexts in Wikipedia and then choose the $N$ distinct types and their $j$ contexts uniformly at random.

\paragraph{Evaluation}
To evaluate models' ability to lemmatize wordforms in their sentence context we follow \citet{bergmanis.goldwater:context} and use the full UDT development and test sets.
Unlike \citet{bergmanis.goldwater:context} who reported token level lemmatization exact match accuracy, we report \textit{type-level} \underline{macro averaged} lemmatization exact match accuracy. This measure better reflects improvements on unseen words, which tend to be rare but are more important (since a most-frequent-lemma baseline does very well on seen words, as shown by \citet{bergmanis.goldwater:context}).

We separately report performance on unseen and ambiguous tokens. For a fair comparison across scenarios with different training sets, we count as unseen only words that are not ambiguous and are absent from \textit{all} training sets/scenarios introduced in Section \ref{sec:results}. 
Due to the small training sets, between 70-90\% of dev set types are classed as unseen in each language.
We define a type as ambiguous if the empirical entropy over its lemmas is greater than 0.1 in the full original UDT training splits.\footnote{This measure, \emph{adjusted ambiguity}, was defined by \citet{Kirefu2018}, who noticed that many frequent wordforms appear to have multiple lemmas due to annotation errors. The adjusted ambiguity filters out these cases.} 
According to this measure, 
only 1.2-5.3\% of dev set types are classed as ambiguous in each language.

\paragraph{Significance Testing}
All systems are trained and tested on ten languages. To test for statistically significant differences between the results of two systems we use a Monte Carlo method: for each set of results (i.e. a set of 10 numerical values) we generate 10000 random samples, where each sample swaps the results of the two systems for each language with a probability of $0.5$. We then obtain a p-value as the proportion of samples for which the difference on average was at least as large as the difference observed in our experiments.

\section{Experiments, Results, and Discussion}\label{sec:results}
\begin{table}[t]

\centering
\small{
\begin{tabular}{crccc}
\toprule
                        &               & \textbf{Ambig.} & \textbf{Unseen} & \textbf{All}    \\ 
                        \midrule
\parbox[t]{2mm}{\multirow{5}{*}{\rotatebox[origin=c]{90}{Tokens}}} & \small\textit{Baseline}      & \cellcolor[HTML]{E9D6CD}41.0 & 26.6 & 31.0 \\
                        & \small\textit{Lemming}       & \cellcolor[HTML]{EBEFEC}38.2 & \cellcolor[HTML]{AFCB87}48.3 & \cellcolor[HTML]{BF71F6}50.6 \\
                        & \small\textit{HMAM}          & \cellcolor[HTML]{E9D6CD}41.4 & \cellcolor[HTML]{92B954}50.2 & \cellcolor[HTML]{A933FC}52.1 \\
                        & \small\textit{Lematus 0-ch}  & \cellcolor[HTML]{EBEFEC}39.9 & \cellcolor[HTML]{DCE6D2}43.7 & \cellcolor[HTML]{D5B0F1}46.8 \\
                        &\small\textit{Lematus 20-ch} & \cellcolor[HTML]{EBEFEC}38.4 & \cellcolor[HTML]{DCE6D2}42.8 & \cellcolor[HTML]{E0CFEE}45.8 \\ \midrule
\parbox[t]{2mm}{\multirow{5}{*}{\rotatebox[origin=c]{90}{Types}}} & \small\textit{Baseline}      & \cellcolor[HTML]{E25D36}45.0 & 26.6 & 32.4 \\
                        & \small\textit{Lemming}       & N/A    & N/A    & N/A    \\
                        & \small\textit{HMAM}          & \cellcolor[HTML]{E6A691}41.8 & \cellcolor[HTML]{669F09}53.7 & \cellcolor[HTML]{9E14FF}56.3 \\
                        & \small\textit{Lematus 0-ch}  & \cellcolor[HTML]{E6A699}42.5 & \cellcolor[HTML]{74a722}53.7 & \cellcolor[HTML]{A933FC}55.1 \\
                        &\small\textit{ Lematus 20-ch} & \cellcolor[HTML]{E25D30}43.1 & \cellcolor[HTML]{74a722}51.7 & \cellcolor[HTML]{A933FC}54.9
\end{tabular}
   \caption{Average type level lemmatization exact match accuracy on five development languages in type and token based training data scenarios. Colour-scale is computed over the whole \textit{Ambig.} column and over all but \textit{Baseline} rows for the other columns.\vspace{-8pt}   \label{table:types_vs_tokens}}}
\end{table}

\begin{figure}
  \centering
  \includegraphics[width=1\linewidth]{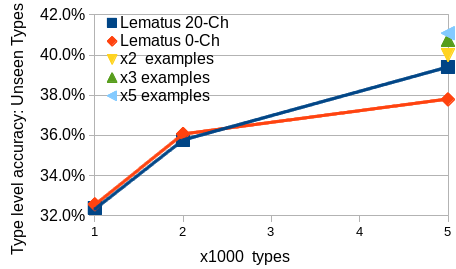}
   \caption{Average type level lemmatization exact match accuracy on unseen words of five development languages. X-axis: thousands of types in training data.\vspace{-8pt}}
  \label{fig:UM_results}
 \end{figure}
 
\begin{table}[t]
\centering
\small{
\begin{tabular}{rllll}
\toprule
                     & \multicolumn{3}{c}{\textbf{DEVELOPMENT}} & \textbf{TEST}   \\
                         \midrule
               & \rot{60}{\textbf{Ambig.}}    &  \rot{60}{\textbf{Unseen}} &\rot{60}{\textbf{All}} & \rot{60}{\textbf{All}}\\
\midrule                
& \multicolumn{4}{c}{\textbf{1k UDT types (No augmentation)}} \\
\midrule                
 \small\textit{Baseline}      & \cellcolor[HTML]{E69D86}\textbf{49.1}    & 30.8   & 36.7   & -  \\
 \small\textit{HMAM}          & \cellcolor[HTML]{E9CEC3}46.3 & \cellcolor[HTML]{C9DBB3}\textbf{58.9}$^{\dagger\ddagger}$   & \cellcolor[HTML]{D7B8F0}\textbf{61.5}$^{\dagger\ddagger}$   & \cellcolor[HTML]{F1C8A4}\textbf{62.6}$^{\dagger\ddagger}$ \\
  \small\textit{Lematus 0-ch}  & \cellcolor[HTML]{E9CEC3}46.5    & \cellcolor[HTML]{DAE5CF}55.0   & \cellcolor[HTML]{EBEFEC}58.5   & \cellcolor[HTML]{ECE5DA}59.1$^{\ddagger} $\\
 \small\textit{Lematus 20-ch} & \cellcolor[HTML]{EBEFEC}45.0    & \cellcolor[HTML]{EBEFEC}54.3   & \cellcolor[HTML]{EBEFEC}57.7   & \cellcolor[HTML]{EBEFEC}57.7 \\
\midrule
& \multicolumn{4}{c}{\textbf{1k UDT types + 1k UM types }} \\
\midrule   
 \small\textit{Baseline}      & \cellcolor[HTML]{EADED7}45.9    & 30.8   & 38.4   & -      \\
 \small\textit{AE Aug Baseline}&\cellcolor[HTML]{EADED7}45.6    & \cellcolor[HTML]{C9DBB3}57.5   & \cellcolor[HTML]{D7B8F0}60.4  & \cellcolor[HTML]{ECE5DA}60.8 \\
  \small\textit{HMAM}          & \cellcolor[HTML]{EADED7}45.9    & \cellcolor[HTML]{B9D196}60.2   & \cellcolor[HTML]{C481F5}64.2   & \cellcolor[HTML]{F5B480}64.3 \\
 \small\textit{Lematus 0-ch } & \cellcolor[HTML]{E9CEC3}46.6    & \cellcolor[HTML]{C9DBB3}59.0   & \cellcolor[HTML]{CA93F3}63.4   & \cellcolor[HTML]{F3BE92}63.6 \\
 \small\textit{Lematus 20-ch }& \cellcolor[HTML]{E69D86}\textbf{49.8}$^*$  &  \cellcolor[HTML]{ABC77A}\textbf{61.7}$^{*\dagger} $ & \cellcolor[HTML]{BE6FF7}\textbf{65.5}$^{*\dagger} $ & \cellcolor[HTML]{F6AA6E}\textbf{65.3}$^{\dagger}$\\
\midrule
 & \multicolumn{4}{c}{\textbf{1k UDT types + 5k UM types }} \\
\midrule   
 \small\textit{Baseline}      & \cellcolor[HTML]{E25D36}\textbf{55.4}$^{*\dagger\ddagger}$   & 30.7   & 41.7   & -      \\
 \small\textit{AE Aug Baseline}&\cellcolor[HTML]{E9CEC3}46.0    & \cellcolor[HTML]{C9DBB3}58.8   & \cellcolor[HTML]{D7B8F0}61.3  &  \cellcolor[HTML]{F1C8A4}61.6\\
  \small\textit{HMAM}          & \cellcolor[HTML]{E9CEC3}46.7    & \cellcolor[HTML]{B9D196}60.8   & \cellcolor[HTML]{BE6FF7}65.7   & \cellcolor[HTML]{F6AA6E}65.7 \\
 \small\textit{Lematus 0-ch}  & \cellcolor[HTML]{E9CEC3}46.2    & \cellcolor[HTML]{ABC77A}61.5   & \cellcolor[HTML]{B75DF8}66.1   & \cellcolor[HTML]{F8A15C}66.4 \\
 \small\textit{Lematus 20-ch} & \cellcolor[HTML]{E7AE9B}48.6    & \cellcolor[HTML]{76A925}\textbf{65.4}$^{*\dagger}$   & \cellcolor[HTML]{A426FD}\textbf{69.2}$^{*\dagger}$   & \cellcolor[HTML]{FB8D38}\textbf{69.5 }$^{*\dagger}$\\
\midrule
& \multicolumn{4}{c}{\textbf{1k UDT types + 10k UM types }} \\
\midrule   
   \small\textit{Baseline}      & \cellcolor[HTML]{E25D36}\textbf{54.9}$^{*\dagger}$   & 31.2   & 43.5   & -      \\
  \small\textit{AE Aug Baseline}&\cellcolor[HTML]{E9CEC3}46.3    & \cellcolor[HTML]{C9DBB3}58.6   & \cellcolor[HTML]{D7B8F0}61.2  & \cellcolor[HTML]{F1C8A4}61.7 \\
 \small\textit{HMAM}          & \cellcolor[HTML]{EADED7}45.4    &  \cellcolor[HTML]{B9D196}60.8   & \cellcolor[HTML]{BE6FF7}65.5   & \cellcolor[HTML]{F6AA6E}65.3 \\
 \small\textit{Lematus 0-ch}  & \cellcolor[HTML]{EADED7}45.5    & \cellcolor[HTML]{97BD5E}62.1   & \cellcolor[HTML]{B14AFA}66.4   & \cellcolor[HTML]{FA974A}66.4 \\
 \small\textit{Lematus 20-ch }& \cellcolor[HTML]{E69D86}49.5$^*$    & \cellcolor[HTML]{669F09}\textbf{66.7}$^{*\dagger}$   & \cellcolor[HTML]{9E14FF}\textbf{70.6}$^{*\dagger}$  & \cellcolor[HTML]{FF7A14}\textbf{70.9}$^{*\dagger}$
\end{tabular}}
   \caption{Average lemmatization accuracy for all 10 languages, trained on 1k UDT types (No aug.), or 1k UDT plus 1k, 5k, or 10k UM types with contexts from Wikipedia. The numerically highest scores in each data setting are bold; $^{*}$, $^{\dagger}$, and $^{\ddagger}$ indicate statistically significant improvements over HMAM \cite{makarov2018neural}, Lematus 0-ch and 20-ch, respectively (all $p<0.05$; see text for details). Colour-scale is computed over the whole \textit{Ambig.} column and over all but \textit{Baseline} rows for the other columns. \vspace{-8pt}}
   \label{table:results_da}
\end{table}
\paragraph{Types vs. Tokens and Context in Very Low Resource Settings}
We compare training on the first 1k tokens vs. first 1k distinct types of the UDT training sets.
Table~\ref{table:types_vs_tokens} shows that if only 1k examples are available, using types is clearly better for all systems. 
Although Lematus does relatively poorly on the token data, it benefits the most from switching to types, putting it on par with HMAM and suggesting is it likely to benefit more from additional type data. 
Lemming requires token-based data, but does worse than HMAM (a context-free method) in the token-based setting, and we also see no benefit from context in comparing Lematus 20-ch vs Lematus 0-ch. So overall, in this very low-resource scenario with no data augmentation, context does not appear to help.\vspace{-5pt}

\paragraph{Using UM + Wikipedia Only}
We now try training only on UM + Wikipedia examples, rather than examples from UDT.
We use 1k, 2k or 5k unambiguous types from UM with a single example context from Wikipedia for each. With 5k types we also try adding more example contexts (2, 3, or 5 examples for each type).

Figure \ref{fig:UM_results} presents the results (for unseen words only). As with the UDT experiments, there is little difference between Lematus 20-ch and Lematus 0-ch in the smallest data setting. However, when the number of training types increases to 5k, the benefits of context begin to show, with Lematus 20-ch yielding a 1.6\% statistically significant ($p < 0.001$) improvement over Lematus 0-ch. The results for increasing the number of examples per type are numerically higher than the one-example case, but the differences are not statistically significant.

It is worth noting that the accuracy even with 5k UM types is considerably lower than the accuracy of the model trained on only 1k UDT types (see Table~\ref{table:types_vs_tokens}). We believe this discrepancy is due to the issues of biased/incomplete data noted above. For example, we analyzed the Latvian data and found that the available tables for nouns, verbs, and adjectives give rise to 78 paradigm slots. The 17 POS tags in UDT give rise to about 10 times as many paradigm slots, although only 448 are present in the unseen words of the dev set. Of these, 197 are represented amongst the 1k UDT training types, whereas only 25 are included in the 1k UM training types. As a result, about $72\%$ of the unseen types of dev set have no representative of their paradigm slot in 1k types of UM, whereas this figure is only $17\%$ for the 1k types of UDT.\vspace{-6pt}

\paragraph{Data Augmentation}
Although UM + Wikipedia examples alone are not sufficient to train a good lemmatizer, they might improve a low-resource baseline trained on UDT data. To see, we augmented the 1k UDT types with  1k, 5k or 10k UM types with contexts from Wikipedia. 

Table~\ref{table:results_da}  summarizes the results, showing that despite the lower quality of the UM + Wikipedia examples, using them improves results of all systems, and more so with more examples. Improvements are especially strong for unseen types, which constitute more than 70\% of types in the dev set.
Furthermore, the benefit of the additional UM examples is above and beyond the effect of auto-encoding (AE Aug Baseline) for all systems in all data scenarios. 

Considering the two context-free models, HMAM does better on the un-augmented 1k UDT data, but (as predicted by our results above) it benefits less from data augmentation than does Lematus 0-ch, so with added data they are statistically equivalent ($p = 0.07$ on the test set with 10k UM).

More importantly, Lematus 20-ch begins to outperform the context-free models with as few as 1k UM + Wikipedia examples, and the difference increases with more examples, eventually reaching over 4\% better on the test set than the next best model (Lematus 0-ch) when 10k UM + Wikipedia examples are used ($p < 0.001$)
This indicates that the system can learn useful contextual cues even from unambiguous training examples.

\begin{figure}[t]
   \includegraphics[width=1\linewidth]{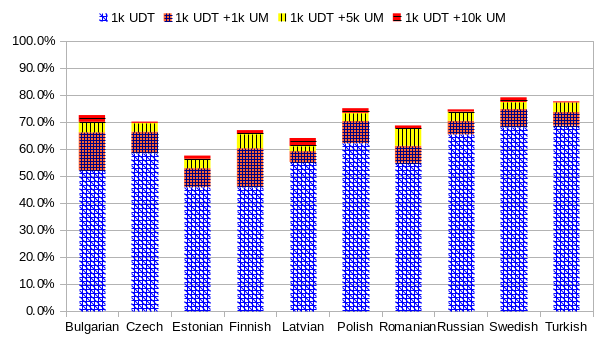}
   \caption{Lematus 20-ch lemmatization accuracy for each language on all types in the dev sets.\vspace{-8pt}}
   \label{fig:individual_languages}
\end{figure}

Finally, Figure \ref{fig:individual_languages} gives a breakdown of Lematus 20-ch dev set accuracy for individual languages, showing that data augmentation helps consistently, although results suggest diminishing returns.

\begin{table}
\centering
\small{

\begin{tabular}{rccc}
\toprule
\textbf{Type accuracy:}   &\textbf{ Ambig.} & \textbf{Unseen} &\textbf{ All}  \\
\midrule
\multicolumn{1}{l}{~~1k UDT+10k UM} & 49.5   & 66.7   & 70.6 \\
\multicolumn{1}{l}{10k UDT tok.}         & 59.6   & 71.4  & 76.6 \\
\multicolumn{1}{l}{10k UDT tok.+10k UM} & \textbf{60.8} & \textbf{75.1} & \textbf{80.1} \\
\midrule
\textbf{Token accuracy:}  & \textbf{Ambig.} & \textbf{Uns.} & \textbf{All}  \\
\midrule
\multicolumn{1}{l}{~~1k UDT+10k UM} & 55.5   & 66.5   & 77.0 \\
\multicolumn{1}{l}{10k UDT tok.}       & \textbf{72.4}   & 72.5    & 85.3 \\
\multicolumn{1}{l}{10k UDT tok.+10k UM} & 72.3 & \textbf{75.3}& \textbf{87.3} \\
\end{tabular}
}
\caption{Lematus 20-ch average lemmatization type and token accuracy for all 10 languages, trained on 1k UDT types, 1k UDT augmented with 10k UM types, 10k UDT continuous tokens, or 10k UDT continuous tokens augmented with 10k UM types. Unless specified otherwise data consists of distinct types.\vspace{-12pt}}
   \label{table:agumentation_vs_udt}
\end{table}
\vspace{-3pt}
\paragraph{Data Augmentation in Medium Resource Setting}
To examine the extent to which augmented data can help in the medium resource setting of 10k continuous tokens of UDT used in previous work, we follow \citet{bergmanis.goldwater:context} and train Lematus 20-ch models for all ten languages using the first 10k tokens of UDT and compare them with models trained on 10k tokens of UDT augmented with 10k UM types.
To provide a better comparison of our results, we report both the type and the token level development set accuracy. 
First of all, Table~\ref{table:agumentation_vs_udt} shows that training on 10k continuous tokens of UDT yields a token level accuracy that is about 8\% higher than when using the 1k types of UDT augmented with 10k UM types---the best-performing data augmentation systems (see Table~\ref{table:results_da}).
Again, we believe this performance gap is due to the issues with the biased/incomplete data noted above.
For example, we analyzed errors that were unique to the model trained on the Latvian augmented data and found that 41\% of the errors were due to wrongly lemmatized words other than nouns, verbs, and adjectives---the three POSs with available inflection tables in UM. For instance, improperly lemmatized pronouns amounted to 14\% of the errors on the Latvian dev set. Table~\ref{table:agumentation_vs_udt} also shows that UM examples with Wikipedia contexts benefit lemmatization not only in the low but also the medium resource setting, yielding statistically significant type and token level accuracy gains over models trained on 10k UDT continuous tokens alone (for both Unseen and All $p < 0.001$).
\vspace{-12pt}
\section{Conclusion}\vspace{-5pt}
We proposed a training data augmentation method that combines the efficiency of type-based learning and the expressive power of a context-sensitive lemmatization model. 
The proposed method uses Wikipedia sentences to provide contextualized examples for  unambiguous inflection-lemma pairs from UniMorph tables.
These examples are noisy and biased, but nevertheless they improve lemmatization accuracy on all ten languages both in low (1k) and medium (10k) resource settings. In particular, we showed that context is helpful, both overall and especially on unseen words---the first work we know of to demonstrate improvements from context in a very low-resource setting.

\bibliography{naaclhlt2019}

\begin{thebibliography}{21}
\expandafter\ifx\csname natexlab\endcsname\relax\def\natexlab#1{#1}\fi

\bibitem[{Nivre~et al.(2017)}]{UDT2}
Joakim Nivre~et al. 2017.
\newblock Universal dependencies 2.0 – {CoNLL} 2017 shared task development
  and test data.
\newblock {LINDAT}/{CLARIN} digital library at the Institute of Formal and
  Applied Linguistics, Charles University.

\bibitem[{Bergmanis and Goldwater(2018)}]{bergmanis.goldwater:context}
Toms Bergmanis and Sharon Goldwater. 2018.
\newblock {Context Sensitive Neural Lemmatization with Lematus}.
\newblock In \emph{Proceedings of the Conference of the North American Chapter
  of the Association for Computational Linguistics: Human Language
  Technologies}.

\bibitem[{Bergmanis et~al.(2017)Bergmanis, Kann, Sch\"{u}tze, and
  Goldwater}]{bergmanis-2017}
Toms Bergmanis, Katharina Kann, Hinrich Sch\"{u}tze, and Sharon Goldwater.
  2017.
\newblock {Training Data Augmentation for Low-Resource Morphological
  Inflection}.
\newblock In \emph{Proceedings of the CoNLL-SIGMORPHON 2017 Shared Task:
  Universal Morphological Reinflection}, Vancouver, Canada. Association for
  Computational Linguistics.

\bibitem[{Chakrabarty et~al.(2017)Chakrabarty, Pandit, and
  Garain}]{chakrabarty2017context}
Abhisek Chakrabarty, Onkar~Arun Pandit, and Utpal Garain. 2017.
\newblock {Context Sensitive Lemmatization Using Two Successive Bidirectional
  Gated Recurrent Networks}.
\newblock In \emph{Proceedings of the 55th Annual Meeting of the Association
  for Computational Linguistics (Volume 1: Long Papers)}, pages 1481--1491,
  Vancouver, Canada. Association for Computational Linguistics.

\bibitem[{Chrupa\l{}a et~al.(2008)Chrupa\l{}a, Dinu, and van
  Genabith}]{chrupala2008learning}
Grzegorz Chrupa\l{}a, Georgiana Dinu, and Josef van Genabith. 2008.
\newblock {Learning Morphology with Morfette}.
\newblock In \emph{Proceedings of the Sixth International Conference on
  Language Resources and Evaluation (LREC'08)}, Marrakech, Morocco. European
  Language Resources Association (ELRA).

\bibitem[{Cotterell et~al.(2016)Cotterell, Kirov, Sylak-Glassman, Yarowsky,
  Eisner, and Hulden}]{cotterell2016sigmorphon}
Ryan Cotterell, Christo Kirov, John Sylak-Glassman, David Yarowsky, Jason
  Eisner, and Mans Hulden. 2016.
\newblock The {SIGMORPHON} 2016 shared task—morphological reinflection.
\newblock In \emph{Proceedings of the 14th {SIGMORPHON Workshop on
  Computational Research in Phonetics, Phonology, and Morphology}}, pages
  10--22.

\bibitem[{Dreyer et~al.(2008)Dreyer, Smith, and Eisner}]{dreyer2008latent}
Markus Dreyer, Jason~R Smith, and Jason Eisner. 2008.
\newblock {Latent-Variable Modeling of String Transductions with Finite-State
  Methods}.
\newblock In \emph{Proceedings of the conference on empirical methods in
  natural language processing}, pages 1080--1089. Association for Computational
  Linguistics.

\bibitem[{Garrette et~al.(2013)Garrette, Mielens, and
  Baldridge}]{garrette2013real}
Dan Garrette, Jason Mielens, and Jason Baldridge. 2013.
\newblock Real-world semi-supervised learning of pos-taggers for low-resource
  languages.
\newblock In \emph{Proceedings of the 51st Annual Meeting of the Association
  for Computational Linguistics (Volume 1: Long Papers)}, volume~1, pages
  583--592.

\bibitem[{Kann and Sch{\"u}tze(2016)}]{kann2016med}
Katharina Kann and Hinrich Sch{\"u}tze. 2016.
\newblock {MED}: The {LMU} system for the {SIGMORPHON} 2016 shared task on
  morphological reinflection.
\newblock In \emph{Proceedings of {ACL}}. Association for Computational
  Linguistics.

\bibitem[{Kann and Sch{\"u}tze(2017)}]{kann2017unlabeled}
Katharina Kann and Hinrich Sch{\"u}tze. 2017.
\newblock {Unlabeled Data for Morphological Generation With Character-Based
  Sequence-to-Sequence Models}.
\newblock In \emph{Proceedings of the First Workshop on Subword and Character
  Level Models in NLP}, pages 76--81.

\bibitem[{Kirefu(2018)}]{Kirefu2018}
Faheem Kirefu. 2018.
\newblock \emph{{Exploring Context Representations for Neural Lemmatisation}}.
\newblock Master's thesis, University of Edinburgh.

\bibitem[{Koehn et~al.(2007)Koehn, Hoang, Birch, Callison-Burch, Federico,
  Bertoldi, Cowan, Shen, Moran, Zens et~al.}]{koehn2007moses}
Philipp Koehn, Hieu Hoang, Alexandra Birch, Chris Callison-Burch, Marcello
  Federico, Nicola Bertoldi, Brooke Cowan, Wade Shen, Christine Moran, Richard
  Zens, et~al. 2007.
\newblock Moses: Open source toolkit for statistical machine translation.
\newblock In \emph{Proceedings of the 45th annual meeting of the association
  for computational linguistics companion volume proceedings of the demo and
  poster sessions}, pages 177--180.

\bibitem[{Makarov and Clematide(2018{\natexlab{a}})}]{makarov2018imitation}
Peter Makarov and Simon Clematide. 2018{\natexlab{a}}.
\newblock {Imitation Learning for Neural Morphological String Transduction}.
\newblock In \emph{Proceedings of the 2018 Conference on Empirical Methods in
  Natural Language Processing}, pages 2877--2882.

\bibitem[{Makarov and Clematide(2018{\natexlab{b}})}]{makarov2018neural}
Peter Makarov and Simon Clematide. 2018{\natexlab{b}}.
\newblock {Neural Transition-based String Transduction for Limited-Resource
  Setting in Morphology}.
\newblock In \emph{Proceedings of the 27th International Conference on
  Computational Linguistics}, pages 83--93.

\bibitem[{McCarthy et~al.(2018)McCarthy, Silfverberg, Cotterell, Hulden, and
  Yarowsky}]{mccarthy2018marrying}
Arya~D McCarthy, Miikka Silfverberg, Ryan Cotterell, Mans Hulden, and David
  Yarowsky. 2018.
\newblock {Marrying Universal Dependencies and Universal Morphology}.
\newblock In \emph{Proceedings of the Second Workshop on Universal Dependencies
  (UDW 2018)}, pages 91--101.

\bibitem[{M\"{u}ller et~al.(2015)M\"{u}ller, Cotterell, Fraser, and
  Sch\"{u}tze}]{muller2015joint}
Thomas M\"{u}ller, Ryan Cotterell, Alexander Fraser, and Hinrich Sch\"{u}tze.
  2015.
\newblock {Joint Lemmatization and Morphological Tagging with Lemming}.
\newblock In \emph{Proceedings of the 2015 Conference on Empirical Methods in
  Natural Language Processing}, pages 2268--2274, Lisbon, Portugal. Association
  for Computational Linguistics.

\bibitem[{Prechelt(1998)}]{prechelt1998early}
Lutz Prechelt. 1998.
\newblock {Early Stopping-but When?}
\newblock \emph{Neural Networks: Tricks of the trade}, pages 553--553.

\bibitem[{Press and Wolf(2017)}]{press2017using}
Ofir Press and Lior Wolf. 2017.
\newblock {Using the Output Embedding to Improve Language Models}.
\newblock In \emph{Proceedings of the 15th Conference of the European Chapter
  of the Association for Computational Linguistics: Volume 2, Short Papers},
  volume~2, pages 157--163.

\bibitem[{Rastogi et~al.(2016)Rastogi, Cotterell, and
  Eisner}]{rastogi2016weighting}
Pushpendre Rastogi, Ryan Cotterell, and Jason Eisner. 2016.
\newblock {Weighting Finite-State Transductions With Neural Context}.
\newblock In \emph{Proceedings of the 2016 Conference of the North American
  Chapter of the Association for Computational Linguistics: Human Language
  Technologies}, pages 623--633.

\bibitem[{Sennrich et~al.(2017)Sennrich, Firat, Cho, Birch, Haddow, Hitschler,
  Junczys{-}Dowmunt, L{\"{a}}ubli, Barone, Mokry, and
  Nadejde}]{DBLP:journals/corr/SennrichFCBHHJL17}
Rico Sennrich, Orhan Firat, Kyunghyun Cho, Alexandra Birch, Barry Haddow,
  Julian Hitschler, Marcin Junczys{-}Dowmunt, Samuel L{\"{a}}ubli, Antonio
  Valerio~Miceli Barone, Jozef Mokry, and Maria Nadejde. 2017.
\newblock Nematus: a toolkit for neural machine translation.
\newblock \emph{CoRR}, abs/1703.04357.

\bibitem[{Wicentowski(2002)}]{wicentowski2002modeling}
Richard Wicentowski. 2002.
\newblock \emph{Modeling and learning multilingual inflectional morphology in a
  minimally supervised framework}.
\newblock Ph.D. thesis, Johns Hopkins University.

\end{thebibliography}
\bibliographystyle{acl_natbib}

\appendix

\section{Lematus Training}\label{appendix:lemmatus_training}
Lematus is implemented using the \textit{Nematus} machine translation toolkit\footnote{\url{https://github.com/EdinburghNLP/nematus}}.
We use default training parameters of Lematus as specified by \citet{bergmanis.goldwater:context}
except for early stopping with patience \cite{prechelt1998early} which we increase to 20. 
Similar to \citet{bergmanis.goldwater:context} we use the first epochs as a burn-in period, after which  we validate the current model by its lemmatization exact match accuracy on the \textit{first 3k instances} of development set and save this model if it performs better than the previous best model.
We choose a burn-in period of 20 and validation interval of 5 epochs for models that we train on datasets up to 2k instances and a burn-in period of 10 and validation interval of 2 epochs for others.
As we work with considerably smaller datasets than \citet{bergmanis.goldwater:context} we reduce the effective model size and increase the rate of convergence by tying the input embeddings of the encoder, the decoder and the softmax output embeddings \cite{press2017using}.

\section{Data Preparation}
Wikipedia database dumps contain XML structured articles that are formatted using the \textit{wikitext} markup language. To obtain wordforms in their sentence context we 1) use \textit{WikiExtractor}\footnote{\url{https://github.com/attardi/wikiextractor}} to extract plain text from Wikipedia database dumps, followed by scripts from \textit{Moses} statistical machine translation system\footnote{\url{https://github.com/moses-smt/mosesdecoder}} \citep{koehn2007moses} to  2)  split text into sentences (\textit{split-sentences.perl}), and 3) extract separate tokens (\textit{tokenizer.perl}).
Finally, we shuffle the extracted sentences to encourage homogeneous type distribution across the entire text.

\begin{table*}[ht]
\section{Result Breakdown by Language}
\centering
\small{
\begin{minipage}{.45\linewidth}

\begin{tabular}{lrccc}
\toprule
\multicolumn{2}{r}{\textbf{Type accuracy:}}                                  & \textbf{Ambig.} & \textbf{Unseen} & \textbf{All} \\
\midrule
\multirow{5}{*}{\rot{90}{\textbf{Bulgarian}}} & \textit{Baseline}        &\textbf{ 63.5 }           & 39.3          & 45.0         \\
                           & \textit{AE Aug Baseline} & -               & -             & -            \\
                           & \textit{HMAM}            & 50.7            & \textbf{61.0 }         & \textbf{63.5 }        \\
                           & \textit{Lematus 0-ch}    & 45.9            & 51.3          & 55.7         \\
                           & \textit{Lematus 20-ch}   & 41.6            & 47.2          & 52.1         \\
\midrule
\multirow{5}{*}{\rot{90}{\textbf{Czech}}}     & \textit{Baseline}        & 38.1            & 31.2          & 33.0         \\
                           & \textit{AE Aug Baseline} & -               & -             & -            \\
                           & \textit{HMAM}            & \textbf{45.2 }           & \textbf{66.8 }         & \textbf{66.7 }        \\
                           & \textit{Lematus 0-ch}    & 40.7            & 59.9          & 60.1         \\
                           & \textit{Lematus 20-ch}   & 40.1            & 58.3          & 58.6         \\
\midrule
\multirow{5}{*}{\rot{90}{\textbf{Estonian}}}  & \textit{Baseline}        & \textbf{51.0 }           & 24.1          & 32.0         \\
                           & \textit{AE Aug Baseline} & -               & -             & -            \\
                           & \textit{HMAM}            & 39.9            & 41.2          & 46.2         \\
                           & \textit{Lematus 0-ch}    & 38.0            &\textbf{ 42.8 }         & \textbf{47.6 }        \\
                           & \textit{Lematus 20-ch}   & 47.8            & 39.9          & 45.9         \\
\midrule
\multirow{5}{*}{\rot{90}{\textbf{Finnish}}}   & \textit{Baseline}        & 46.4            & 21.3          & 26.1         \\
                           & \textit{AE Aug Baseline} & -               & -             & -            \\
                           & \textit{HMAM}            & \textbf{44.7}            & \textbf{48.0}          & \textbf{50.4 }        \\
                           & \textit{Lematus 0-ch}    & 44.4            & 41.5          & 44.9         \\
                           & \textit{Lematus 20-ch}   & 44.6            & 43.0          & 46.0         \\
\midrule
\multirow{5}{*}{\rot{90}{\textbf{Latvian}}}   & \textit{Baseline}        & 42.4            & 25.6          & 31.6         \\
                           & \textit{AE Aug Baseline} & -               & -             & -            \\
                           & \textit{HMAM}            & 44.0            & \textbf{52.6}          &\textbf{ 55.6}         \\
                           & \textit{Lematus 0-ch}    & \textbf{47.1}            & 51.8          & 55.2         \\
                           & \textit{Lematus 20-ch}   & 43.1            & 52.1          & 55.2         \\
\midrule
\multirow{5}{*}{\rot{90}{\textbf{Polish}}}    & \textit{Baseline}        & \textbf{42.9}            & 26.6          & 33.3         \\
                           & \textit{AE Aug Baseline} & -               & -             & -            \\
                           & \textit{HMAM}            & 41.2            & \textbf{60.5 }         & 62.4         \\
                           & \textit{Lematus 0-ch}    & 40.9            & 60.4          & \textbf{62.6}         \\
                           & \textit{Lematus 20-ch}   & 35.5            & 59.7          & 62.2         \\
\midrule
\multirow{5}{*}{\rot{90}{\textbf{Romanian}}}  & \textit{Baseline}        & 27.6            & 34.9          & 40.0         \\
                           & \textit{AE Aug Baseline} & -               & -             & -            \\
                           & \textit{HMAM}            & 38.8            & \textbf{55.1}          & \textbf{57.9 }        \\
                           & \textit{Lematus 0-ch}    & \textbf{44.6 }           & 50.2          & 54.5         \\
                           & \textit{Lematus 20-ch}   & 40.7            & 50.9          & 54.9         \\
\midrule
\multirow{5}{*}{\rot{90}{\textbf{Russian}}}   & \textit{Baseline}        & 43.0            & 34.9          & 39.0         \\
                           & \textit{AE Aug Baseline} & -               & -             & -            \\
                           & \textit{HMAM}            & 39.3            & \textbf{66.4 }         & \textbf{67.0}         \\
                           & \textit{Lematus 0-ch}    & 42.3            & 63.4          & 65.4         \\
                           & \textit{Lematus 20-ch}   & \textbf{44.6 }           & 63.7          & 65.5         \\
\midrule
\multirow{5}{*}{\rot{90}{\textbf{Swedish}}}   & \textit{Baseline}        & 77.8            & 42.8          & 52.7         \\
                           & \textit{AE Aug Baseline} & -               & -             & -            \\
                           & \textit{HMAM}            & 58.5            & \textbf{67.7}          & \textbf{72.6}         \\
                           & \textit{Lematus 0-ch}    & \textbf{59.5 }           & 64.1          & 70.1         \\
                           & \textit{Lematus 20-ch}   & 54.0            & 62.6          & 68.1         \\
\midrule
\multirow{5}{*}{\rot{90}{\textbf{Turkish}}}   & \textit{Baseline}        & 58.8            & 26.6          & 33.6         \\
                           & \textit{AE Aug Baseline} & -               & -             & -            \\
                           & \textit{HMAM}            & 60.2            & \textbf{69.6 }         & \textbf{72.3}         \\
                           & \textit{Lematus 0-ch}    & \textbf{61.8}            & 64.7          & 68.4         \\
                           & \textit{Lematus 20-ch}   & 58.2            & 65.4          & 68.6        
\end{tabular}
\caption{Individual type level  lemmatization accuracy for all 10 languages on development set, trained on 1k~UDT~types (no augmentation) with contexts from Wikipedia. The numerically highest scores for each language are bold. For the summary of results see Table~\ref{table:results_da}.}
\end{minipage}
\hspace{0.45cm}
\begin{minipage}{.45\linewidth}



\begin{tabular}{lrccc}
\toprule
\textbf{}                           & \textbf{Type accuracy:}  & \textbf{Ambig.} & \textbf{Unseen} & \textbf{All}  \\
\midrule
\multirow{5}{*}{\rot{90}{\textbf{Bulgarian}}} & \textit{Baseline}        & \textbf{64.3}   & 39.3          & 47.2          \\
                                    & \textit{AE Aug Baseline} & 49.2            & 60.3          & 63.3          \\
                                    & \textit{HMAM}            & 41.4            & \textbf{63.8} & \textbf{67.4} \\
                                    & \textit{Lematus 0-ch}    & 49.2            & 59.2          & 64.2          \\
                                    & \textit{Lematus 20-ch}   & 53.3            & 61.7          & 66.2          \\
                                    \midrule
\multirow{5}{*}{\rot{90}{\textbf{Czech}}}     & \textit{Baseline}        & 40.3            & 31.2          & 34.2          \\
                                    & \textit{AE Aug Baseline} & 42.5            & 63.6          & 63.6          \\
                                    & \textit{HMAM}            & 42.5            & \textbf{64.9} & \textbf{66.6} \\
                                    & \textit{Lematus 0-ch}    & 38.9            & 58.7          & 60.9          \\
                                    & \textit{Lematus 20-ch}   & \textbf{53.3}   & 61.7          & 66.2 \\
                                    \midrule
\multirow{5}{*}{\rot{90}{\textbf{Estonian}}}  & \textit{Baseline}        & \textbf{58.1}   & 24.1          & 34.4          \\
                                    & \textit{AE Aug Baseline} & 41.4            & 42.6          & 47.4          \\
                                    & \textit{HMAM}            & 47.9            & 43.8          & 51.4          \\
                                    & \textit{Lematus 0-ch}    & 48.1            & 45.2          & 52.5          \\
                                    & \textit{Lematus 20-ch}   & 45.4            & \textbf{46.3} & \textbf{52.9} \\
                                    \midrule
\multirow{5}{*}{\rot{90}{\textbf{Finnish}}}   & \textit{Baseline}        & 46.3            & 21.3          & 27.4          \\
                                    & \textit{AE Aug Baseline} & 44.3            & 43.3          & 45.5          \\
                                    & \textit{HMAM}            & 45.6            & 55.7          & 59.3          \\
                                    & \textit{Lematus 0-ch}    & 47.1            & 55.2          & 59.3          \\
                                    & \textit{Lematus 20-ch}   & \textbf{49.2}   & \textbf{56.6} & \textbf{60.3} \\
                                    \midrule
\multirow{5}{*}{\rot{90}{\textbf{Latvian}}}   & \textit{Baseline}        & 45.4            & 25.6          & 33.7          \\
                                    & \textit{AE Aug Baseline} & 38.8            & 52.2          & 54.9          \\
                                    & \textit{HMAM}            & 42.7            & 52.7          & 56.9          \\
                                    & \textit{Lematus 0-ch}    & 45.2            & 51.8          & 56.4          \\
                                    & \textit{Lematus 20-ch}   & \textbf{48.6}   & \textbf{56.3} & \textbf{59.2} \\
                                    \midrule
\multirow{5}{*}{\rot{90}{\textbf{Polish}}}    & \textit{Baseline}        & \textbf{46.3}   & 26.6          & 35.4          \\
                                    & \textit{AE Aug Baseline} & 37.5            & 62.7          & 64.8          \\
                                    & \textit{HMAM}            & 37.4            & 62.0          & 66.4          \\
                                    & \textit{Lematus 0-ch}    & 45.3            & 62.3          & 67.1          \\
                                    & \textit{Lematus 20-ch}   & 38.1            & \textbf{66.9} & \textbf{70.3} \\
                                    \midrule
\multirow{5}{*}{\rot{90}{\textbf{Romanian}}}  & \textit{Baseline}        & 37.7            & 34.9          & 43.3          \\
                                    & \textit{AE Aug Baseline} & 42.6            & 53.2          & 57.0          \\
                                    & \textit{HMAM}            & 48.3            & \textbf{57.3} & 62.7          \\
                                    & \textit{Lematus 0-ch}    & \textbf{51.1}   & 57.0          & \textbf{62.8} \\
                                    & \textit{Lematus 20-ch}   & 49.0            & 55.9          & 61.2          \\
                                    \midrule
\multirow{5}{*}{\rot{90}{\textbf{Russian}}}   & \textit{Baseline}        & 44.4            & 34.7          & 40.6          \\
                                    & \textit{AE Aug Baseline} & 42.6            & 65.7          & 67.1          \\
                                    & \textit{HMAM}            & 43.2            & 66.1          & 68.3          \\
                                    & \textit{Lematus 0-ch}    & 38.7            & 64.6          & 67.3          \\
                                    & \textit{Lematus 20-ch}   & \textbf{50.3}   & \textbf{67.6} & \textbf{70.4} \\
                                    \midrule
\multirow{5}{*}{\rot{90}{\textbf{Swedish}}}   & \textit{Baseline}        & \textbf{78.4}   & 42.8          & 54.2          \\
                                    & \textit{AE Aug Baseline} & 58.4            & 64.8          & 70.3          \\
                                    & \textit{HMAM}            & 53.5            & 68.9          & 73.8          \\
                                    & \textit{Lematus 0-ch}    & 48.8            & \textbf{70.9} & \textbf{75.4} \\
                                    & \textit{Lematus 20-ch}   & 56.4            & 69.7          & 74.8          \\
                                    \midrule
\multirow{5}{*}{\rot{90}{\textbf{Turkish}}}   & \textit{Baseline}        & 59.9            & 26.6          & 35.5          \\
                                    & \textit{AE Aug Baseline} & 58.6            & 66.9          & 69.9          \\
                                    & \textit{HMAM}            & 56.1            & 67.0          & 69.4          \\
                                    & \textit{Lematus 0-ch}    & 54.1            & 65.2          & 67.8          \\
                                    & \textit{Lematus 20-ch}   & \textbf{62.9}   & \textbf{70.6} & \textbf{73.7}
\end{tabular}
\caption{Individual type level  lemmatization accuracy for all 10 languages on development set, trained on 1k~UDT~types plus 1k UM types with contexts from Wikipedia. The numerically highest scores for each language are bold. For the summary of results see Table~\ref{table:results_da}.}
\end{minipage}}
\end{table*}

\begin{table*}[ht]
\centering
\small{
\begin{minipage}{.45\linewidth}

\begin{tabular}{lrccc}
\toprule
\multicolumn{2}{r}{\textbf{Type accuracy:}}                    & \textbf{Ambig.} & \textbf{Unseen}   & \textbf{All}    \\
\midrule
\multirow{5}{*}{\rot{90}{\textbf{Bulgarian}}} & \textit{Baseline}        & \textbf{67.2\%} & 39.3\%          & 50.0\%          \\
                                    & \textit{AE Aug Baseline} & 47.9\%          & 62.6\%          & 65.0\%          \\
                                    & \textit{HMAM}            & 44.3\%          & \textbf{68.2\%} & \textbf{72.1\%} \\
                                    & \textit{Lematus 0-ch}    & 43.1\%          & 67.0\%          & 71.1\%          \\
                                    & \textit{Lematus 20-ch}   & 50.4\%          & 65.9\%          & 70.0\%          \\
                                    \midrule
\multirow{5}{*}{\rot{90}{\textbf{Czech}}}     & \textit{Baseline}        & 43.0\%          & 31.2\%          & 36.8\%          \\
                                    & \textit{AE Aug Baseline} & 43.2\%          & 66.9\%          & 66.6\%          \\
                                    & \textit{HMAM}            & 41.0\%          & 61.9\%          & 64.7\%          \\
                                    & \textit{Lematus 0-ch}    & 39.5\%          & 61.6\%          & 64.4\%          \\
                                    & \textit{Lematus 20-ch}   & \textbf{42.6\%} & \textbf{68.4\%} & \textbf{69.7\%} \\
                                    \midrule
\multirow{5}{*}{\rot{90}{\textbf{Estonian}}}  & \textit{Baseline}        & 62.9\%          & 24.1\%          & 37.1\%          \\
                                    & \textit{AE Aug Baseline} & 43.1\%          & 40.3\%          & 45.3\%          \\
                                    & \textit{HMAM}            & 48.0\%          & 44.9\%          & 53.3\%          \\
                                    & \textit{Lematus 0-ch}    & 51.3\%          & 45.2\%          & 53.5\%          \\
                                    & \textit{Lematus 20-ch}   & \textbf{48.6\%} & \textbf{49.7\%} & \textbf{56.3\%} \\
                                    \midrule
\multirow{5}{*}{\rot{90}{\textbf{Finnish}}}   & \textit{Baseline}        & 49.4\%          & 21.3\%          & 30.3\%          \\
                                    & \textit{AE Aug Baseline} & 42.5\%          & 44.9\%          & 47.6\%          \\
                                    & \textit{HMAM}            & 44.0\%          & 58.4\%          & 62.5\%          \\
                                    & \textit{Lematus 0-ch}    & 45.9\%          & 60.8\%          & 64.7\%          \\
                                    & \textit{Lematus 20-ch}   & \textbf{52.5\%} & \textbf{61.9\%} & \textbf{65.5\%} \\
                                    \midrule
\multirow{5}{*}{\rot{90}{\textbf{Latvian}}}   & \textit{Baseline}        & \textbf{45.6\%} & 25.6\%          & 35.9\%          \\
                                    & \textit{AE Aug Baseline} & 39.6\%          & 53.4\%          & 55.3\%          \\
                                    & \textit{HMAM}            & 45.2\%          & 52.3\%          & 57.6\%          \\
                                    & \textit{Lematus 0-ch}    & 43.8\%          & 54.5\%          & 59.1\%          \\
                                    & \textit{Lematus 20-ch}   & 44.7\%          & \textbf{57.6\%} & \textbf{61.1\%} \\
                                    \midrule
\multirow{5}{*}{\rot{90}{\textbf{Polish}} }   & \textit{Baseline}        & \textbf{50.4\%} & 26.6\%          & 39.2\%          \\
                                    & \textit{AE Aug Baseline} & 38.8\%          & 64.1\%          & 66.2\%          \\
                                    & \textit{HMAM}            & 41.6\%          & 62.3\%          & 68.4\%          \\
                                    & \textit{Lematus 0-ch}    & 43.3\%          & 65.2\%          & 70.7\%          \\
                                    & \textit{Lematus 20-ch}   & 40.3\%          & \textbf{69.7\%} & \textbf{73.4\%} \\
                                    \midrule
\multirow{5}{*}{\rot{90}{\textbf{Romanian}}}  & \textit{Baseline}        & 44.3\%          & 34.9\%          & 47.9\%          \\
                                    & \textit{AE Aug Baseline} & 41.3\%          & 54.9\%          & 58.4\%          \\
                                    & \textit{HMAM}            & 50.2\%          & 58.4\%          & 65.6\%          \\
                                    & \textit{Lematus 0-ch}    & 51.4\%          & 60.8\%          & 67.2\%          \\
                                    & \textit{Lematus 20-ch}   & \textbf{47.9\%} & \textbf{62.6\%} & \textbf{67.7\%} \\
                                    \midrule
\multirow{5}{*}{\rot{90}{\textbf{Russian}}}   & \textit{Baseline}        & 48.5\%          & 34.7\%          & 44.4\%          \\
                                    & \textit{AE Aug Baseline} & 42.1\%          & 65.5\%          & 66.5\%          \\
                                    & \textit{HMAM}            & 46.4\%          & 65.5\%          & 69.7\%          \\
                                    & \textit{Lematus 0-ch}    & 40.5\%          & 64.4\%          & 68.5\%          \\
                                    & \textit{Lematus 20-ch}   & \textbf{42.7\%} & \textbf{71.1\%} & \textbf{73.8\%} \\
                                    \midrule
\multirow{5}{*}{\rot{90}{\textbf{Swedish}}}   & \textit{Baseline}        & \textbf{80.6\%} & 42.8\%          & 58.0\%          \\
                                    & \textit{AE Aug Baseline} & 58.7\%          & 67.3\%          & 71.4\%          \\
                                    & \textit{HMAM}            & 51.9\%          & \textbf{72.6\%} & \textbf{77.7\%} \\
                                    & \textit{Lematus 0-ch}    & 49.1\%          & 71.4\%          & 76.2\%          \\
                                    & \textit{Lematus 20-ch}   & 49.5\%          & 72.3\%          & 77.4\%          \\
                                    \midrule
\multirow{5}{*}{\rot{90}{\textbf{Turkish}}}   & \textit{Baseline}        & 61.8\%          & 26.6\%          & 37.9\%          \\
                                    & \textit{AE Aug Baseline} & 62.8\%          & 68.5\%          & 71.2\%          \\
                                    & \textit{HMAM}            & 54.2\%          & 63.6\%          & 65.7\%          \\
                                    & \textit{Lematus 0-ch}    & 53.6\%          & 63.9\%          & 65.5\%          \\
                                    & \textit{Lematus 20-ch}   & \textbf{67.1\%} & \textbf{74.6\%} & \textbf{77.2\%}
\end{tabular}
\caption{Individual type level  lemmatization accuracy for all 10 languages on development set, trained on 1k~UDT~types plus 5k UM types with contexts from Wikipedia. The numerically highest scores for each language are bold. For the summary of results see Table~\ref{table:results_da}.}

\end{minipage}
\hspace{0.45cm}
\begin{minipage}{.45\linewidth}

\begin{tabular}{rrccc}
\toprule
\multicolumn{2}{r}{\textbf{Type accuracy:}}                                  & \textbf{Ambig.}        & \textbf{Unseen }         &\textbf{ All}           \\
\midrule
\multirow{5}{*}{\rot{90}{\textbf{Bulgarian}}} & \textit{Baseline}        & \textbf{66.2} & 39.7          & 51.2          \\
                           & \textit{AE Aug Baseline} & 48.1          & 62.8          & 65.4          \\
                           & \textit{HMAM}            & 42.7          & \textbf{70.6} & \textbf{74.2} \\
                           & \textit{Lematus 0-ch}    & 44.8          & 67.4          & 71.4          \\
                           & \textit{Lematus 20-ch}   & 50.6          & 68.4          & 72.5          \\
                           \midrule
\multirow{5}{*}{\rot{90}{\textbf{Czech}}}     & \textit{Baseline}        & 43.2          & 31.5          & 38.2          \\
                           & \textit{AE Aug Baseline} & 44.9          & 68.0          & 68.1          \\
                           & \textit{HMAM}            & 41.1          & 61.7          & 64.7          \\
                           & \textit{Lematus 0-ch}    & 38.1          & 61.9          & 64.6          \\
                           & \textit{Lematus 20-ch}   & \textbf{42.7} & \textbf{68.7} & \textbf{70.1} \\
\midrule
\multirow{5}{*}{\rot{90}{\textbf{Estonian}}}  & \textit{Baseline}        & \textbf{62.0} & 24.3          & 37.8          \\
                           & \textit{AE Aug Baseline} & 45.8          & 41.0          & 45.8          \\
                           & \textit{HMAM}            & 48.9          & 45.4          & 53.7          \\
                           & \textit{Lematus 0-ch}    & 46.6          & 45.7          & 53.8          \\
                           & \textit{Lematus 20-ch}   & 44.9          & \textbf{51.3} & \textbf{57.5} \\
\midrule
\multirow{5}{*}{\rot{90}{\textbf{Finnish}}}   & \textit{Baseline}        & 49.3          & 21.9          & 32.7          \\
                           & \textit{AE Aug Baseline} & 41.6          & 46.0          & 48.5          \\
                           & \textit{HMAM}            & 45.0          & 59.2          & 62.7          \\
                           & \textit{Lematus 0-ch}    & 43.2          & 62.8          & 66.2          \\
                           & \textit{Lematus 20-ch}   & \textbf{49.4} & \textbf{63.8} & \textbf{67.0} \\
\midrule
\multirow{5}{*}{\rot{90}{\textbf{Latvian}}}   & \textit{Baseline}        & 46.7          & 25.8          & 36.9          \\
                           & \textit{AE Aug Baseline} & 41.6          & 52.0          & 54.6          \\
                           & \textit{HMAM}            & 44.6          & 53.8          & 59.0          \\
                           & \textit{Lematus 0-ch}    & 42.6          & 55.3          & 59.7          \\
                           & \textit{Lematus 20-ch}   & \textbf{47.7} & \textbf{60.6} & \textbf{64.0} \\
\midrule
\multirow{5}{*}{\rot{90}{\textbf{Polish}}}    & \textit{Baseline}        & \textbf{48.7} & 27.0          & 42.1          \\
                           & \textit{AE Aug Baseline} & 36.8          & 64.3          & 65.4          \\
                           & \textit{HMAM}            & 44.0          & 60.9          & 66.4          \\
                           & \textit{Lematus 0-ch}    & 46.2          & 67.2          & 72.4          \\
                           & \textit{Lematus 20-ch}   & 42.0          & \textbf{71.2} & \textbf{75.1} \\
\midrule
\multirow{5}{*}{\rot{90}{\textbf{Romanian}}}  & \textit{Baseline}        & 43.7          & 35.5          & 49.6          \\
                           & \textit{AE Aug Baseline} & 41.0         & 54.3          & 57.2          \\
                           & \textit{HMAM}            & 45.6          & 56.8          & 63.7          \\
                           & \textit{Lematus 0-ch}    & \textbf{50.3} & 61.7          & 67.8          \\
                           & \textit{Lematus 20-ch}   & 49.5          & \textbf{63.4} & \textbf{68.7} \\
\midrule
\multirow{5}{*}{\rot{90}{\textbf{Russian}}}   & \textit{Baseline}        & \textbf{50.2} & 35.4          & 47.1          \\
                           & \textit{AE Aug Baseline} & 46.8          & 65.6          & 67.0          \\
                           & \textit{HMAM}            & 39.1          & 64.6          & 68.2          \\
                           & \textit{Lematus 0-ch}    & 38.7          & 64.6          & 67.3          \\
                           & \textit{Lematus 20-ch}   & 47.3          & \textbf{71.2} & \textbf{74.7} \\
\midrule
\multirow{5}{*}{\rot{90}{\textbf{Swedish}}}   & \textit{Baseline}        & \textbf{77.3} & 43.0          & 59.9          \\
                           & \textit{AE Aug Baseline} & 58.8          & 66.9          & 71.7          \\
                           & \textit{HMAM}            & 47.3          & 73.0          & 77.7          \\
                           & \textit{Lematus 0-ch}    & 55.5          & 74.1          & 78.6          \\
                           & \textit{Lematus 20-ch}   & 55.6          & \textbf{75.1} & \textbf{79.1} \\
\midrule
\multirow{5}{*}{\rot{90}{\textbf{Turkish}}}   & \textit{Baseline}        & 62.4          & 27.1          & 39.5          \\
                           & \textit{AE Aug Baseline} & 57.9          & 67.7          & 70.3          \\
                           & \textit{HMAM}            & 55.9          & 62.4          & 64.8          \\
                           & \textit{Lematus 0-ch}    & 49.1          & 60.7          & 62.5          \\
                           & \textit{Lematus 20-ch}   & \textbf{65.8} & \textbf{73.6} & \textbf{76.8}
\end{tabular}
\caption{Individual type level lemmatization accuracy for all 10 languages on development set, trained on 1k~UDT~types plus 10k UM types with contexts from Wikipedia. The numerically highest scores for each language are bold.  For the summary of results see Table~\ref{table:results_da}.}
\end{minipage}}
\end{table*}
\begin{table*}
\centering
\small

\begin{tabular}{llcccccc}
\toprule
                           &                                            & \multicolumn{3}{c}{\textbf{Type level accuracy:}}    & \multicolumn{3}{c}{\textbf{Token level accuracy:}}   \\
                           & \multicolumn{1}{c}{\textbf{Training data}} & \textbf{Ambig.} & \textbf{Unseen}   & \textbf{All}    & \textbf{Ambig.} & \textbf{Unseen}   & \textbf{All}    \\
\midrule
\multirow{2}{*}{\textbf{Bulgarian}} & 10k UDT tok.                               & \textbf{62.3} & 75.7          & 80.1          & 72.3          & 75.7          & 89.5          \\
                           & 10k UDT tok. + 10k UM types                & 62.2          & \textbf{78.7} & \textbf{83.6} & \textbf{73.3} & \textbf{78.1} & \textbf{91.0} \\
\midrule
\multirow{2}{*}{\textbf{Czech}}     & 10k UDT tok.                               & 49.7          & 76.4          & 77.8          & \textbf{80.7} & 77.7          & 88.3          \\
                           & 10k UDT tok. + 10k UM types                & \textbf{52.4} & \textbf{78.3} & \textbf{80.4} & 80.0          & \textbf{80.0} & \textbf{89.6} \\
\midrule
\multirow{2}{*}{\textbf{Estonian}}  & 10k UDT tok.                               & 65.3          & 54.0          & 64.5          & 80.1          & 54.3          & 76.8          \\
                           & 10k UDT tok. + 10k UM types                & \textbf{65.9} & \textbf{63.4} & \textbf{72.6} & \textbf{81.5} & \textbf{64.2} & \textbf{82.4} \\
                           \midrule
\multirow{2}{*}{\textbf{Finnish}}   & 10k UDT tok.                               & \textbf{60.7} & 60.1          & 66.5          & \textbf{73.8} & 62.4          & 78.2          \\
                           & 10k UDT tok. + 10k UM types                & 57.8          & \textbf{63.7} & \textbf{69.4} & 70.3          & \textbf{66.0} & \textbf{79.8} \\
                           \midrule
\multirow{2}{*}{\textbf{Latvian}}   & 10k UDT tok.                               & 57.5          & 70.9          & 75.6          & 69.2          & 70.5          & 82.6          \\
                           & 10k UDT tok. + 10k UM types                & \textbf{58.9} & \textbf{73.6} & \textbf{77.8} & \textbf{70.2} & \textbf{73.8} & \textbf{84.4} \\
                           \midrule
\multirow{2}{*}{\textbf{Polish}}    & 10k UDT tok.                               & \textbf{59.8} & 78.7          & 83.6          & \textbf{76.5} & 78.8          & 89.5          \\
                           & 10k UDT tok. + 10k UM types                & 57.4          & \textbf{81.2} & \textbf{86.1} & 71.3          & \textbf{81.4} & \textbf{90.9} \\
                           \midrule
\multirow{2}{*}{\textbf{Romanian}}  & 10k UDT tok.                               & 51.7          & 61.1          & 66.6          & 54.1          & 60.6          & 79.1          \\
                           & 10k UDT tok. + 10k UM types                & \textbf{57.1} & \textbf{68.2} & \textbf{74.2} & \textbf{60.7} & \textbf{68.2} & \textbf{83.9} \\
                           \midrule
\multirow{2}{*}{\textbf{Russian}}   & 10k UDT tok.                               & \textbf{64.4} & 80.5          & 83.5          & \textbf{65.9} & 80.8          & 88.5          \\
                           & 10k UDT tok. + 10k UM types                & 61.1          & \textbf{82.6} & \textbf{85.9} & 59.9          & \textbf{82.7} & \textbf{89.8} \\
                           \midrule
\multirow{2}{*}{\textbf{Swedish}}   & 10k UDT tok.                               & 63.2          & 74.9          & 80.9          & 78.5          & 73.6          & 89.6          \\
                           & 10k UDT tok. + 10k UM types                & \textbf{65.1} & \textbf{78.4} & \textbf{83.7} & \textbf{79.0} & \textbf{75.9} & \textbf{90.4} \\
                           \midrule
\multirow{2}{*}{\textbf{Turkish}}   & 10k UDT tok.                               & 64.2          & 82.1          & 87.1          & 73.1          & 81.8          & 91.2 \\
                           & 10k UDT tok. + 10k UM types                & 69.9          & \textbf{82.9} & \textbf{87.3} & \textbf{76.9} & \textbf{82.7} & \textbf{91.5}
\end{tabular}
\caption{Individual type and token level lemmatization accuracy for all 10 languages on development set for Lematus~20-ch models trained on 10k~UDT tokens and 10k~UDT tokens plus 10k UM types with contexts from Wikipedia. The numerically highest scores for each language are bold. For the summary of results see Table~\ref{table:agumentation_vs_udt}.}
\end{table*}
\FloatBarrier

\end{document}